\documentclass[letterpaper, 10 pt, conference]{ieeeconf}  % Comment this line out if you need a4paper

\IEEEoverridecommandlockouts                              % This command is only needed if
\usepackage{amsmath}
\usepackage{amssymb}
\usepackage{graphicx}
\usepackage{url}
\usepackage{placeins}
\usepackage{balance}
\usepackage{bm}
\usepackage{multirow}
\usepackage{booktabs}
\newcommand{\tabstack}[2]{\begin{tabular}[b]{@{}c@{}}#1\\#2\end{tabular}} % two-line table header
\makeatletter
\def\fnum@table{Table~\thetable}
\let\IEEEfloatmakecaption\@makecaption
\long\def\@makecaption#1#2{%
  \ifx\@captype\@IEEEtablestring
    \setbox\@tempboxa\hbox{\footnotesize #1.~~ #2}%
    \ifdim\wd\@tempboxa>\hsize
      \setbox\@tempboxa\hbox{\footnotesize #1.~~ }%
      \parbox[t]{\hsize}{\footnotesize\noindent\unhbox\@tempboxa#2}%
    \else
      \hbox to\hsize{\footnotesize\hfil\box\@tempboxa\hfil}%
    \fi
    \@IEEEtablecaptionsepspace
  \else
    \IEEEfloatmakecaption{#1}{#2}%
  \fi}
\makeatother
\usepackage[
    pagebackref,
    breaklinks,
    colorlinks=black,
    linkcolor=red,
    citecolor=black,
    filecolor=black,
]{hyperref}
\title{Which Trajectory Is Safer? \\ Combining Rule and Learning for Safety Evaluation}

\author{Xuewen Luo$^{1,*}$, Linlian Jiang$^{3}$, Bo Yu$^{1}$, Fan Ding$^{2}$, Xinxin Zuo$^{3}$, and Yang Wang$^{3}$% <-this % stops a space
\thanks{$^{*}$Corresponding author}%
\thanks{$^{1}$Xuewen Luo and Bo Yu are with the University of Utah.}%
\thanks{$^{2}$Fan Ding is with Monash University.}%
\thanks{$^{3}$Linlian Jiang, Xinxin Zuo, and Yang Wang are with Concordia University.}%
}

\begin{document}

\maketitle
\thispagestyle{empty}
\pagestyle{empty}

%%%%%%%%%%%%%%%%%%%%%%%%%%%%%%%%%%%%%%%%%%%%%%%%%%%%%%%%%%%%%%%%%%%%%%%%%%%%%%%%
\begin{abstract}

% Bo Yu 上面整个一句话的中间的部分读起来有点别扭，我下面换了个写完，你可以看看哪个好点。
% Collision rate is the standard metric for trajectory safety, yet it is binary and coarse: once planners avoid collisions in nearly all cases, it saturates and offers no signal about how safe a collision-free trajectory actually is.
%Collision rate remains the dominant metric for evaluating the safety of autonomous driving trajectories, yet as planners increasingly avoid collisions outright, it reveals little about the safety differences among trajectories that remain collision-free. We introduce \textbf{FutureRisk}, a metric that quantifies the \emph{latent future risk} of an executed trajectory from its own realized interactions with surrounding agents, whether logged or rolled out in closed-loop simulation. The resulting score is continuous and separates collision-free trajectories by how closely and how persistently they approached an unsafe interaction state.At its core, a \textbf{Context-Adaptive Risk Model (CARM)} replaces fixed, hand-crafted TTC-style thresholds with a learned, context-dependent safety criterion, while remaining bounded around an interpretable, closed-form geometric prior.  Experiments show that FutureRisk agrees with native collision labels more closely than existing surrogate safety measures while keeping most of its score anchored to that prior, transfers zero-shot across datasets, and orders collision-free trajectories that collision rate cannot distinguish, exposing planner behavior that a binary outcome would report as safe.

Collision rate remains a dominant metric for evaluating autonomous
driving trajectory safety, but its binary nature obscures substantial
safety differences among collision-free trajectories. Existing
surrogate-safety metrics provide finer-grained assessments, yet often
trade off physical interpretability against adaptivity to interaction
context. We introduce \textbf{FutureRisk}, a continuous metric that
quantifies the \emph{latent future risk} of a trajectory by how closely
and persistently its realized interactions approach unsafe states. At
its core, our \textbf{Context-Adaptive Risk Model (CARM)} combines a
context-dependent safety-time distribution with an independently
supervised critical-event-time distribution. Its safety criterion is bounded around a closed-form physical prior, while a bounded residual provides additional adaptation without abandoning this
physical anchor. Experiments on DeepAccident and Bench2Drive show that
FutureRisk achieves stronger agreement with native collision labels than existing surrogate-safety measures, discriminates risk among collision-free trajectories, and transfers zero-shot across datasets. Comparisons across four planners further show that FutureRisk reveals
safety differences hidden by binary collision outcomes. Together, these
results establish FutureRisk as a fine-grained, interpretable, and
context-adaptive metric for trajectory safety evaluation. Code is available at \url{https://github.com/Xuewen2025/safety-evaluation}.

\end{abstract}

% We introduce \textbf{FutureRisk}, a metric that scores the \emph{latent future risk} of a trajectory from its own realized future interactions, whether logged from human driving or rolled out by a policy in closed-loop simulation, producing a continuous score that distinguishes collision-free trajectories by how closely and how persistently they approached an unsafe interaction state. 

% The resulting score is continuous, interpretable, and directly comparable across trajectories on matched scenarios, enabling fine-grained safety evaluation beyond binary collision outcomes. 

% bo yu
% 感觉上面的这部分可以改成，Experiment results show that the proposed latent future risk and its corresponding CARM can achieve .... 从实验结果出发比较好，而不是重复上面的论述，因为ocontinuous, intepretable 已经在上面出现论述过了。用实验结果去论证上面提到过的比较好

%最后可以加一句升华：By exposing near-miss behavior that collision rate alone would deem safe, FutureRisk helps identify hazardous planning behavior before it reaches public roads, supporting the safer deployment of autonomous driving systems. 当然也可以更简略一点最后一句
% \end{abstract}

%%%%%%%%%%%%%%%%%%%%%%%%%%%%%%%%%%%%%%%%%%%%%%%%%%%%%%%%%%%%%%%%%%%%%%%%%%%%%%%%

\section{Introduction}

Recent planning systems for autonomous driving have achieved substantial
improvements in collision avoidance, resulting in increasingly low
observed collision rates~\cite{hamidaoui2025survey}. Nevertheless,
collision occurrence, typically summarized by the collision rate, remains
a dominant safety measure. As a binary outcome, it reveals little about
safety differences among trajectories that ultimately remain
collision-free~\cite{weng2020mprism} and may yield overly optimistic
assessments by hiding residual risks that never turn into a
collision~\cite{jiang2025interactive,cui2025unified}. In practice, a
collision-free trajectory is not necessarily sufficiently safe: some trajectories maintain
ample safety margins throughout an interaction, whereas others enter
safety-critical states with little room for recovery or avoid contact
only because surrounding agents respond favorably. Treating these
outcomes as equally safe obscures substantial differences in their safety
profiles and motivates a more fine-grained, continuous evaluation measure.

\begin{figure}[t]
    \centering    \includegraphics[width=\columnwidth]
    {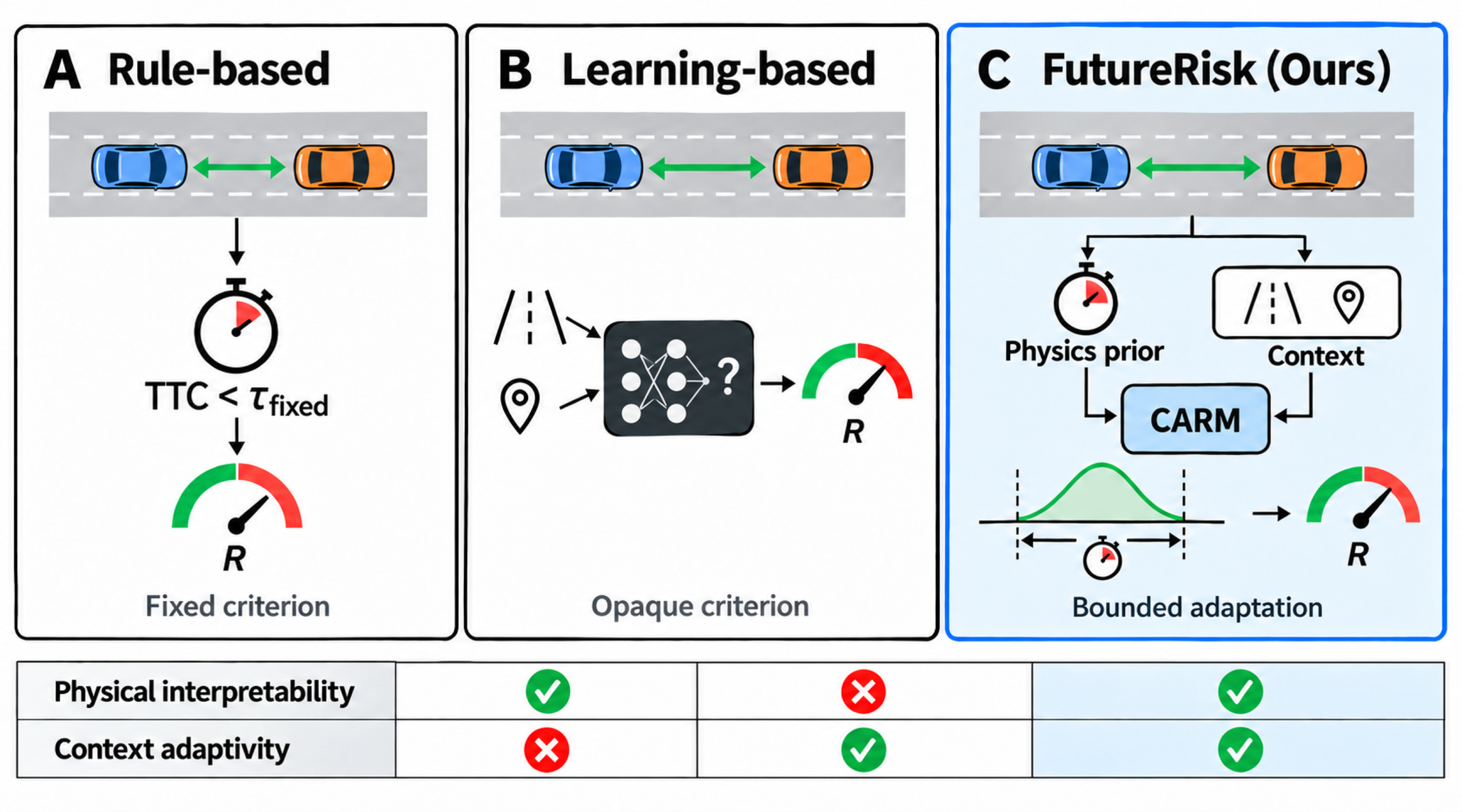}
    \caption{Comparison of trajectory-safety metrics. 
    (A) Rule-based metrics are physically interpretable but rely on fixed
    criteria. (B) Learning-based metrics adapt to context but lack an
    explicit physical anchor. (C) FutureRisk combines physical
    interpretability with bounded, context-adaptive learning.}
    \vspace{-2mm}
    \label{fig:futurerisk-method-comparison}
\end{figure}

To move beyond binary collision outcomes, prior work has moved toward surrogate-safety and criticality criteria that quantify proximity to hazardous interactions \cite{arun2021systematic}. Among formula-based criteria, Time-to-Collision (TTC) provides closed-form temporal reasoning, but becomes a safety
judgment only when compared with a required safety time. Conventional TTC-based criteria usually fix that requirement using scenario-agnostic
assumptions and thresholds that may not reflect the margin needed in different driving contexts \cite{westhofen2023criticality}. Learning-based
alternatives can adapt to context but often lack a physical structure for interpreting their outputs \cite{lu2022learning,jiao2026gssm}. Recent hybrid
approaches combine learned context modeling with physical or rule-based safety structures \cite{tian2025cri,lu2025piram,liao2025safecast,yang2026vlmcase}, but target safety-constrained control, motion forecasting, or context-conditioned collision-risk estimation rather than continuous
trajectory-level evaluation. Consequently, as Fig.~\ref{fig:futurerisk-method-comparison} illustrates, a gap remains for a safety metric that is both physically interpretable and adaptive to interaction context.

% xuewen
%Together, these gaps motivate us to reconsider how the safety of an executed trajectory should be characterized beyond its eventual collision outcome. A more informative safety metric should not only distinguish different levels of risk among collision-free trajectories, but also remain interpretable while adapting its safety criterion to the interaction context. Drawing on prior notions of prospective safety and criticality \cite{wang2021review}, we characterize risk at each point along an executed trajectory by how closely and how persistently its subsequent evolution approaches an unsafe interaction state. This risk is \emph{future} with respect to the evaluated point, but is computed retrospectively from the future evolution that actually occurred. It is \emph{latent} because it may remain hidden when the interaction does not ultimately result in a collision. We therefore refer to this quantity as \textbf{latent future risk}. Quantifying latent future risk places collision-free trajectories along a continuous spectrum of safety, addressing the limitations of binary collision outcomes; doing so through an interpretable yet context-adaptive criterion further resolves the trade-off exhibited by existing criticality measures. This leads to the central question of this work: \textbf{How can the latent future risk of an executed trajectory be quantified in a manner that is both interpretable and adaptive to scenario context?}

%final

The binary nature of collision outcomes motivates a more informative notion of trajectory safety. We characterize risk by how closely and how
persistently a trajectory approaches an unsafe interaction state over its subsequent realized evolution. We call this quantity \textbf{latent future
risk}: \emph{future} because it is evaluated from the interaction evolution that follows each point on the trajectory, and \emph{latent} because it may
never materialize as a collision. Quantifying latent future risk places collision-free trajectories on a continuous safety spectrum rather than under one binary label. The remaining challenge is to quantify this risk in a manner that retains physical interpretability while adapting to the
interaction context. This leads to our central question: \textbf{How can latent future risk be quantified by a metric that is both physically
interpretable and adaptive to scenario context?}

% These gaps motivate a more informative metric for evaluating trajectory safety. Two trajectories may both avoid collision while differing substantially in safety: one may maintain a comfortable margin, whereas the other may pass dangerously close to another agent. We therefore evaluate not only whether a collision occurs, but also how closely the trajectory approaches an unsafe interaction state. We call this quantity \textbf{latent future risk}: it is \emph{future} because it reflects the subsequent evolution from each evaluated point, and \emph{latent} because the underlying risk may not ultimately materialize as a collision. By quantifying this risk, collision-free trajectories can be placed on a continuous safety spectrum rather than assigned the same binary outcome. This leads to our central question: \textbf{How can latent future risk be quantified using a metric that is both interpretable and adaptive to scenario context?}

To this end, we propose \textbf{FutureRisk}, a continuous metric that quantifies the latent future risk of a trajectory from its interactions with surrounding agents. At its core, our \textbf{Context-Adaptive Risk Model (CARM)} combines an independently supervised critical-event-time distribution with a context-dependent safety-time distribution. The median of the safety-time distribution is quantitatively bounded around a closed-form physical prior derived from reaction time, closing speed, available deceleration, and prediction uncertainty, while a bounded logit residual provides additional calibration without allowing unrestricted deviation from the structured estimate. 
FutureRisk aggregates the resulting per-agent risks into a trajectory-level score, revealing fine-grained safety differences among trajectories that share the same collision outcome.
Because it operates directly on trajectories without requiring planner-specific internal representations, FutureRisk supports both individual-trajectory auditing and comparisons across planning systems and data sources.
% Because it operates directly on trajectories without requiring planner-specific internal representations, FutureRisk supports both individual-trajectory auditing and comparisons across planning systems and data sources. 
Experiments on DeepAccident and Bench2Drive evaluate its agreement with collision outcomes, discrimination among collision-free trajectories, zero-shot cross-dataset transfer, and application to four representative planning methods.
% Experiments on DeepAccident and Bench2Drive evaluate its agreement with collision outcomes, discrimination among collision-free trajectories, zero-shot cross-dataset transfer, and application to four representative planning methods.

Our main contributions are summarized as follows:

\begin{itemize}

\item We introduce \textbf{FutureRisk}, a continuous trajectory-level safety metric that scores latent future risk from realized interactions and
distinguishes trajectories with the same collision outcome.

\item We develop \textbf{CARM}, which combines a context-adaptive safety-time
distribution, its median bounded within a factor of two of a closed-form
physics prior, with an independently supervised critical-event
distribution and a bounded logit residual.

\item We evaluate FutureRisk on DeepAccident~\cite{deepaccident} and Bench2Drive~\cite{bench2drive}, examining collision agreement, discrimination among collision-free trajectories, and zero-shot transfer. Comparisons across four planning methods and qualitative analyses further assess its utility for trajectory safety evaluation.

\end{itemize}

\section{Related Work}
\label{sec:related}

\noindent\textbf{Beyond collision rate.}
Collision occurrence records a safety failure but gives no resolution
among trajectories that remain collision-free. Surrogate safety measures
therefore assess proximity to hazardous interactions and evasive behavior
before a crash occurs~\cite{arun2021systematic,wang2021review}, and MPrISM
argues that observing eventual collisions is insufficient to judge whether
a system stays in a continuously safe state~\cite{weng2020mprism}. The
coarseness persists in current benchmarks: nuPlan's closed-loop score
blends progress, comfort, and minimum TTC into a continuous value, yet
gates it to zero on any at-fault collision, so its safety judgment
collapses to pass/fail the moment a violation
occurs~\cite{caesar2021nuplan}, and in the Waymo Open Sim Agents
Challenge a static policy drives collision rate to zero without being
safe~\cite{montali2023wosac}. FutureRisk answers this need with a
continuous trajectory-level score.

\noindent\textbf{Rule-based safety criteria.}
TTC, post-encroachment time (PET), and deceleration rate to avoid a crash quantify temporal proximity or required evasive effort, while Responsibility-Sensitive Safety (RSS) specifies explicit safety envelopes \cite{wang2021review,shalevshwartz2017rss}. Their explicit formulations make assumptions and outputs inspectable, but their suitability depends on the interaction geometry, motion model, and parameter choices \cite{westhofen2023criticality}. These limitations are not inherent to all analytical methods: Li \emph{et al.} extend surrogate measures to higher-dimensional motion and richer vehicle dynamics \cite{li2024beyond}. Nevertheless, analytical structure alone does not provide a learned safety criterion that adapts to interaction context.

\noindent\textbf{Data-driven and hybrid safety criteria.}
Lu \emph{et al.} learn representations of TTC and maneuver sequences and identify conflicts through clustering, reducing reliance on manually selected thresholds \cite{lu2022learning}. GSSM learns context-conditioned spacing distributions and derives risk through extreme-value theory \cite{jiao2026gssm}. Other methods combine prediction with structured risk calculations: Interactive Risk integrates multimodal trajectory prediction with driving risk fields, while RMPP quantifies overlap between predicted position distributions \cite{jiang2025interactive,cui2025unified}. These methods demonstrate that learning and interpretable risk structure can coexist. FutureRisk instead emphasizes a specific combination: a context-adaptive safety-time distribution whose median is bounded around a closed-form physical prior, critical-event timing supervised independently of that learned safety-time prediction, and a bounded residual fitted after freezing the structured model.

\section{Method}
\label{sec:method}

\subsection{Overview}
\label{sec:method_overview}

\begin{figure*}[t]
    \vspace{-5mm}
    \centering
    \includegraphics[width=0.8\textwidth]{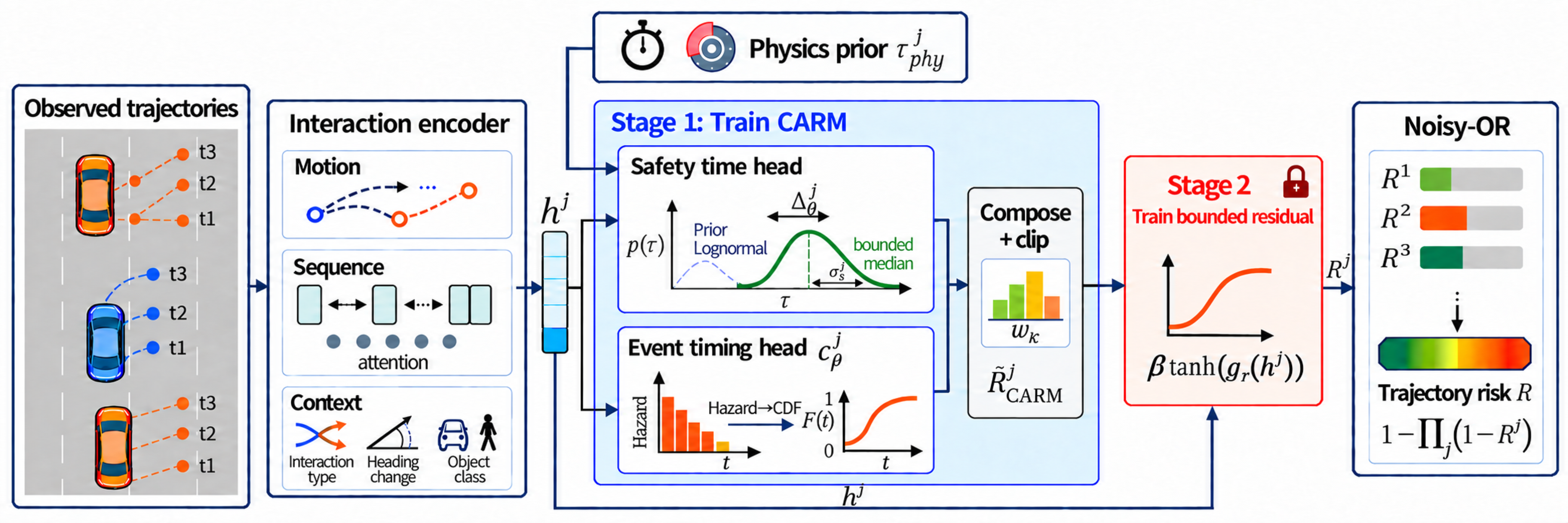}
    \caption{Overview of FutureRisk. A shared encoding $\bm h^{j}$ feeds a
safety-time head bounded around the physics prior
$\tau_{\mathrm{phy}}^{j}$ and an independent event-timing head, composed
into $\widetilde R_{\mathrm{CARM}}^{j}$. Blue is trained in Stage~1, the
red residual in Stage~2 with the rest frozen; noisy-OR gives $R$.}
    \vspace{-1mm}
    \label{fig:futurerisk_method}
\end{figure*}

FutureRisk scores an executed ego trajectory by how close it came to an
unsafe interaction with the agents around it. Its score is
\begin{equation}
R = f_{\mathrm{FR}}\big(\bm{\xi}_e,\mathcal{W}^{\mathrm{obs}}\big)\in[0,1],
\label{eq:future_risk_objective}
\end{equation}
where a larger $R$ means the trajectory approached an unsafe interaction
state more closely or stayed near one for longer. A binary collision label
gives the same value to a trajectory that kept a wide margin and to one
that avoided contact by a fraction of a second; $R$ separates them.
Let $\bm{\xi}_e$ denote the
ego trajectory over a horizon $[0,H]$ and
$\mathcal{W}^{\mathrm{obs}}=\{\bm{\xi}_j\}_{j=1}^{J}$ the trajectories of
the $J$ surrounding agents over the same horizon, both read from what
actually happened; FutureRisk is a retrospective score, not an online
predictor.
Throughout, time is discretized into $K$ bins with edges
$0=b_0<b_1<\dots<b_K=H$, every time-dependent quantity below is evaluated
on this grid, and $\sigma(\cdot)$ denotes the logistic function with
inverse $\operatorname{logit}(\cdot)$.

At the core of $f_{\mathrm{FR}}$ is CARM (Fig.~\ref{fig:futurerisk_method}), a learned risk model whose
safety criterion adapts to context while staying bounded around a
closed-form physics prior. For each agent $j$, CARM forms a structured
estimate $\widetilde R_{\mathrm{CARM}}^{j}$, refines it with a bounded
residual (Sec.~\ref{sec:bounded_residual}), and $f_{\mathrm{FR}}$
aggregates the per-agent risks by noisy-OR
(Sec.~\ref{sec:final_noisy_or}):
\begin{equation}
\begin{aligned}
R&=1-\prod_{j}\left(1-R^{j}\right),\\[2pt]
R^{j}&=\sigma\!\left(
\operatorname{logit}\big(\widetilde R_{\mathrm{CARM}}^{j}\big)
+
\underbrace{\beta\tanh\!\left(g_r(\bm h^{j})\right)}_{\text{learned, Stage 2}}
\right),
\end{aligned}
\label{eq:future_risk_decomposition}
\end{equation}
where $\bm h^{j}$ is a shared encoding of the ego--agent interaction
(Sec.~\ref{sec:CARM_Architecture}), $g_r$ a scalar head, and $\beta$ a
fixed bound. Because $R$ grows with the number of agents, trajectories
are compared only on matched scenarios with the same surrounding agents.

The structured estimate composes two distributions: required safety time
and critical-event time,
\begin{equation}
\widetilde R_{\mathrm{CARM}}^{j}
=
\Gamma\!\Big(
\underbrace{\log\tau_{\mathrm{phy}}^{j}}_{\text{raw}}
+\underbrace{\Delta_{\theta}^{j}}_{\substack{\text{learned,}\\\text{Stage 1}}},\ \
\underbrace{\sigma_{s}^{j}}_{\substack{\text{learned,}\\\text{Stage 1}}},\ \
\underbrace{c_{\theta}^{j}}_{\substack{\text{learned,}\\\text{Stage 1}}}
\Big).
\label{eq:carm_score_ingredients}
\end{equation}
$\tau_{\mathrm{phy}}^{j}$ is the only closed-form quantity: a physics
prior for the required safety margin built from reaction time, closing
speed, and available deceleration. $\Delta_\theta^j$ and $\sigma_s^j$
turn it into a bounded log-normal safety-time distribution; $c_\theta^j$
is the critical-event CDF sequence, independently supervised so that
criticality is never defined through the margin it is combined with.
Sec.~\ref{sec:CARM_Architecture} details each term and $\Gamma$;
Sec.~\ref{sec:carm_training} derives the target that supervises $R$, the
label for $c_\theta^j$, and the two-stage schedule. The geometry-aware
3D-TTC $\tau_{\mathrm{3D}}^{j}$ enters only that target, never CARM's
forward computation.

\subsection{CARM Architecture}
\label{sec:CARM_Architecture}

\subsubsection{Interaction Encoder}

A bidirectional gated recurrent unit (BiGRU) with attention pooling
encodes $\bm{X}_{\mathrm{int}}^{j}$, while separate multilayer
perceptrons (MLPs) encode $\bm{x}_{\mathrm{kin}}^{j}$ and
$\bm{x}_{\mathrm{ctx}}^{j}$. A fusion MLP combines the three outputs,
followed by layer normalization:
\begin{equation}
\bm h^{j}
=f_{\mathrm{enc}}\!\left(
\bm{x}_{\mathrm{kin}}^{j},
\bm{X}_{\mathrm{int}}^{j},
\bm{x}_{\mathrm{ctx}}^{j}
\right).
\label{eq:carm_encoder}
\end{equation}
For ego trajectory $\bm{\xi}_e$ and agent $j$, the encoder inputs are
current relative kinematics $\bm{x}_{\mathrm{kin}}^{j}$ (relative
position and velocity, relative heading, closing speed, and both current
speeds), the observed sequence $\bm{X}_{\mathrm{int}}^{j}$ (relative 3D
motion and signed clearance over $[0,H]$), and semantic context
$\bm{x}_{\mathrm{ctx}}^{j}$ (interaction-type distribution,
relative-heading statistics, ego heading change, and object class).
Signed clearance is the center distance minus the sum of the two boxes' half-diagonals.

\subsubsection{Safety-Time Head Modeling}

The safety-time head uses the physics prior
\begin{equation}
\tau_{\mathrm{phy}}^{j}(t)
=
\mathrm{clip}\!\left(
t_{\mathrm{react}}+\frac{v_{\mathrm{close}}^{j}(t)}{a_{\mathrm{avail}}}+\epsilon_{\mathrm{unc}},
\ [\tau_{\min},\tau_{\max}]
\right),
\label{eq:phy_prior}
\end{equation}
where $\tau_{\mathrm{phy}}^{j}(t)$ is the prior safety time for agent $j$,
$v_{\mathrm{close}}^{j}(t)$ is the closing speed along the current
relative bearing, $t_{\mathrm{react}}$ is the reaction time,
$a_{\mathrm{avail}}$ is the available deceleration, and
$\epsilon_{\mathrm{unc}}$ is an uncertainty pad. The parameters and
clipping range are object-class specific.

The head models the required safety time as
\begin{align}
\Delta_\theta^j
&=\Delta_{\max}\tanh\!\left(g_{\mu}(\bm h^{j})\right),
\nonumber\\
\sigma_s^{j}
&=\sigma_{\min}+(\sigma_{\max}-\sigma_{\min})
\sigma\!\left(g_{\sigma}(\bm h^{j})\right),
\nonumber\\
\mu_{s,k}^{j}
&=\log\tau_{\mathrm{phy}}^{j}(b_k)+\Delta_\theta^j,
\nonumber\\
T_{s,k}^{j}&\sim\operatorname{LogNormal}\!\left(\mu_{s,k}^{j},(\sigma_s^{j})^2\right),
\label{eq:adaptive_safe_ttc}
\end{align}
Here, $\tau_{\mathrm{phy}}^{j}$ provides the bin-wise prior safety time,
$\Delta_\theta^j$ shifts its log-space median, and $\sigma_s^j$ controls
the spread of the resulting log-normal distribution. Thus
$\exp(\mu_{s,k}^{j})=\tau_{\mathrm{phy}}^{j}(b_k)\exp(\Delta_\theta^j)$.
With $\Delta_{\max}=\log 2$, this median lies in
$[\tau_{\mathrm{phy}}^{j}(b_k)/2,\,2\tau_{\mathrm{phy}}^{j}(b_k)]$.
$\tau_{\mathrm{phy}}^{j}$ and $\mu_{s,k}^{j}$ vary across bins, while
$\Delta_\theta^j$ and $\sigma_s^j$ are shared across bins for each agent. Collecting the bin-wise location parameters gives the safety-time output
$\bm{\mu}_s^j=\{\mu_{s,k}^j\}_{k=1}^K$.

\subsubsection{Event Timing Head Modeling}

The event-time head produces the critical-event cumulative distribution function (CDF) sequence $c_\theta^j=\{F_k^j\}_{k=1}^K$ over the same $K$ bins. It first predicts the conditional hazard
$q_k^j=\sigma(g_k^{\mathrm{evt}}(\bm h^j))$, the probability of an event in bin $k$ given no earlier event. The event probability mass and CDF are
\begin{equation}
\begin{aligned}
p_k^j&=q_k^j\prod_{\ell<k}(1-q_\ell^j),
&F_k^j&=\sum_{\ell=1}^{k}p_\ell^j,
\end{aligned}
\label{eq:event_time_distribution}
\end{equation}
where $p_k^j$ is the event probability mass in bin $k$, and $F_k^j$ is
the probability of an event by bin $k$. The remaining mass
$1-F_K^j=\prod_{\ell=1}^{K}(1-q_\ell^j)$ represents no event within the horizon.

\subsubsection{Risk Composition and Clipping}

Using the safety-time outputs $\bm{\mu}_s^j$ and $\sigma_s^j$ and the
event-time output $c_\theta^j$ defined above, the composition module
computes
$\widetilde R_{\mathrm{CARM}}^{j}=\Gamma(\bm{\mu}_s^j,\sigma_s^j,c_\theta^j)$.
Using the standard normal CDF $\Phi$, it computes the safety-time interval
weights
\begin{align}
w_{1\hphantom{K+}}^j
&=\Phi\!\Bigl(\tfrac{\log b_1-\mu_{s,1}^j}{\sigma_s^j}\Bigr),\nonumber\\
w_{k\hphantom{K+}}^j
&=\Phi\!\Bigl(\tfrac{\log b_k-\mu_{s,k}^j}{\sigma_s^j}\Bigr)
-\Phi\!\Bigl(\tfrac{\log b_{k-1}-\mu_{s,k}^j}{\sigma_s^j}\Bigr),
\ k=2,\ldots,K,\nonumber\\
w_{K+1}^j
&=1-\Phi\!\Bigl(\tfrac{\log b_K-\mu_{s,K}^j}{\sigma_s^j}\Bigr).
\end{align}
$w_k^j$ corresponds to interval $(b_{k-1},b_k]$, with $F_{T_s}(0)=0$ used
for $w_1^j$, and $w_{K+1}^j$ to $(b_K,\infty)$. The implementation
lower-bounds values passed to the logarithm by $\epsilon_t=10^{-6}$.
Because each interval uses its local $\mu_{s,k}^j$, the weights are not
renormalized and need not sum to one. $\Gamma$ then combines each weight
with the event CDF:
\begin{align}
R_{\mathrm{CARM}}^{j}
&=\sum_{k=1}^{K}F_k^j w_k^j
+F_K^j w_{K+1}^j,
\label{eq:carm_composition}
\end{align}
where the tail weight $w_{K+1}^j$ is paired with the final CDF value
$F_K^j$. The resulting score is clipped to
\begin{equation}
\widetilde R_{\mathrm{CARM}}^{j}=\mathrm{clip}\!\left(R_{\mathrm{CARM}}^{j},\ \epsilon,\ 1-\epsilon\right),
\label{eq:carm_clip}
\end{equation}
where $\epsilon=10^{-4}$, producing the structured interaction score
$\widetilde R_{\mathrm{CARM}}^{j}$.

\subsection{Bounded Residual Modeling}
\label{sec:bounded_residual}

The residual head maps the shared interaction representation $\bm h^j$
to the bounded logit correction
\begin{equation}
x_\theta^j := \beta\tanh\!\left(g_r(\bm h^{j})\right), \qquad \beta=2.0,
\label{eq:bounded_carm_residual}
\end{equation}
where $g_r$ is a scalar head and $x_\theta^j\in[-\beta,\beta]$. It is
added to the structured logit and mapped back to probability space:
\begin{equation}
R^j=\sigma\!\left(
\operatorname{logit}\big(\widetilde R_{\mathrm{CARM}}^j\big)+x_\theta^j
\right).
\label{eq:bounded_residual_score}
\end{equation}

\subsection{Final Noisy-OR Aggregation}
\label{sec:final_noisy_or}

For the $J$ surrounding agents, the final trajectory-level risk is the
noisy-OR of their refined interaction risks:
\begin{equation}
R=1-\prod_{j=1}^{J}\left(1-R^j\right).
\label{eq:final_noisy_or}
\end{equation}
Each $R^j$ contributes one multiplicative complement, producing the final
FutureRisk score $R\in[0,1]$.

\subsection{Two-Stage Training}
\label{sec:carm_training}

Training has two stages. Stage~1 trains CARM to model structured risk
from collision and near-miss targets; Stage~2 trains the bounded residual
to refine the structured score.

\subsubsection{Collision-Target Modeling}

Geometry-aware 3D time-to-collision is
\begin{equation}
\begin{split}
\tau_{\mathrm{3D}}^{j}(t)
=
\inf\!\{
&\delta\geq0:
\mathcal{B}_{e}(t+\delta)
\cap
\mathcal{B}_{j}(t+\delta)
\neq\varnothing,\;\\
&t+\delta\le H
\},
\end{split}
\label{eq:high_level_3d_ttc}
\end{equation}
Let $\mathcal{B}_{e}(t)$ and $\mathcal{B}_{j}(t)$ denote the oriented 3D
occupied volumes of the ego and agent $j$ at time $t$, read directly from
$\bm{\xi}_e$ and the observed trajectory of agent $j$ in
$\mathcal{W}^{\mathrm{obs}}$: the trajectory's own already-realized
future, not a forward projection from the current state. Here
$\tau_{\mathrm{3D}}^{j}(t)=+\infty$ if no overlap occurs within
$[t,H]$. Because the volumes follow the realized trajectories rather than
forward extrapolation under fixed controls, $\tau_{\mathrm{3D}}^{j}$ requires
no motion assumption or prediction and applies uniformly to rear-end,
lateral, crossing, and head-on interactions.

\begin{equation}
m^{j}(t)
=
\tau_{\mathrm{3D}}^{j}(t)-\tau_{\mathrm{phy}}^{j}(t),
\qquad
m
=
\min_{j}\min_{t\in[0,H]} m^{j}(t),
\label{eq:future_safety_margin}
\end{equation}
Here, $m^j(t)$ is the collision margin for agent $j$, and $m$ is its
minimum over all agents and times. It uses the fixed physics prior
$\tau_{\mathrm{phy}}^j(t)$ from Eq.~\eqref{eq:phy_prior}, without
$\Delta_\theta^j$. Positive and negative margins indicate states outside
and inside the collision boundary, respectively. The margin is mapped to
\begin{equation}
r(t)
=
\phi\!\left(\min_j m^j(t)\right)
=
\sigma\!\left(-\frac{\min_j m^j(t)}{T_m}\right)
\in[0,1],
\label{eq:continuous_risk}
\end{equation}
where $T_m>0$ is the transition width. Its peak is the trajectory-level target,
\begin{equation}
y
=
\max_{t\in[0,H]} r(t)
=
\phi(m).
\label{eq:analytic_risk_target}
\end{equation}
Thus, $y$ is the collision target computed from $\bm{\xi}_e$,
$\mathcal{W}^{\mathrm{obs}}$, and $\tau_{\mathrm{phy}}^{j}$. It also
serves as the analytic-only baseline $y_{\mathrm{analytic}}$ in
Sec.~\ref{sec:experiments}.

By construction, $\tau_{\mathrm{3D}}^j(t)$ is finite only for actual box
overlap; hence $m=+\infty$ and $y=0$ for all non-overlapping trajectories,
including near misses. The near-miss target below provides event-time
supervision for these interactions.

\subsubsection{Near-Miss-Target Modeling}

The critical-event label is built from evasive kinematics and observed
collisions. Let
$\mathrm{sev}(a_{\mathrm{lon}},a_{\mathrm{lat}})=\sqrt{\max(-a_{\mathrm{lon}},0)^2+a_{\mathrm{lat}}^2}$
denote a party's evasive-severity score, where $a_{\mathrm{lon}}$ is the
signed derivative of speed, negative under braking, and
$a_{\mathrm{lat}}$ is lateral acceleration; it is evaluated for the ego
and for agent $j$, taking the larger,
\begin{equation*}
a^{j}(t)=\max\!\big(
\mathrm{sev}(a_{\mathrm{lon}}^{e}(t),a_{\mathrm{lat}}^{e}(t)),\,
\mathrm{sev}(a_{\mathrm{lon}}^{j}(t),a_{\mathrm{lat}}^{j}(t))
\big).
\end{equation*}
A critical event is the first time this signal exceeds a threshold
$\theta_{d,c}$ calibrated for dataset $d$ and interaction type
$c=c(t,j)$, or a collision occurs:
\begin{equation}
T_{\mathrm{evt}}^{j}
=\inf\!\left\{t\in[0,H]:
\mathrm{collision}^{j}(t)
\ \lor\
a^{j}(t)>\theta_{d,c(t,j)}\right\},
\label{eq:carm_critical_event}
\end{equation}
with $T_{\mathrm{evt}}^j=+\infty$ when neither occurs within $H$. Each
$\theta_{d,c}$ is the $95^{\mathrm{th}}$ percentile of $a^{j}(t)$ for
interaction type $c$ within dataset $d$; types below a minimum sample
count are merged into an \emph{other} category. Thresholds are fit
separately per dataset to avoid cross-dataset distortion.

Writing $k_j\in\{1,\dots,K\}$ for the bin containing $T_{\mathrm{evt}}^j$
(bin $K$ for a censored agent with no event inside the horizon), only bins
$k\le k_j$ are at risk. Writing  $e^j=\mathbf{1}[T_{\mathrm{evt}}^j<\infty]$
for whether an event occurred, so $e^j=0$ for a censored agent regardless
of $k_j$,
\begin{align}
\mathcal L_{\mathrm{evt}}
=\,&-\frac{1}{N}\sum_{j}\sum_{k=1}^{k_j}
\Big[
w_e\cdot e^j\mathbf{1}[k{=}k_j]\log q_k^j
\nonumber\\
&+\big(1-e^j\mathbf{1}[k{=}k_j]\big)\log(1-q_k^j)
\Big],
\label{eq:l_evt}
\end{align}
where $N$ counts unmasked agents,
so a censored agent contributes only negative terms across all $K$ bins,
reinforcing low hazard rather than a spurious final-bin event. $w_e$ is
the positive-class weight, the inverse frequency of agents with finite
$T_{\mathrm{evt}}^j$ on the training split.

\subsubsection{Two-Stage Procedure}

The final FutureRisk score is trained to approximate the collision target:
\begin{equation}
R=f_{\mathrm{FR}}(\bm\xi_e,\mathcal W^{\mathrm{obs}})\approx y.
\label{eq:carm_estimator}
\end{equation}
The corresponding risk loss is
\begin{equation}
\mathcal L_{\mathrm{risk}}=\operatorname{BCE}(R,y),
\label{eq:l_risk}
\end{equation}
with a positive-class ($y>0.5$) weight set by inverse frequency on the
training split, as for $w_e$. The safety-time branch has no direct
supervision: $T_{s,k}^j$ has no observable ground truth, so pseudo-labels
for it would be circular. Gradients into
$\mu_{s,k}^j,\sigma_s^j$ thus arrive only through the composition in
Eq.~\eqref{eq:carm_composition}, plus a Stage-1 penalty that keeps the two
unsupervised heads out of saturation,
\begin{equation}
\mathcal L_{\mathrm{sat}}
=\frac{1}{|\mathcal V|}\sum_{j\in\mathcal V}
\left[(g_{\mu}^j)^2+(g_{\sigma}^j)^2\right],
\label{eq:l_sat}
\end{equation}
where $\mathcal V$ is the set of unmasked agents and $g_\mu^j,g_\sigma^j$
are the raw head outputs before $\tanh$ and $\sigma$. Training uses two
stages,
\begin{equation}
\begin{aligned}
\mathcal L_{\mathrm{Stage\,1}}
&=\lambda_e\mathcal L_{\mathrm{evt}}
+\lambda_r\mathcal L_{\mathrm{risk}}
+\lambda_{\mathrm{sat}}\mathcal L_{\mathrm{sat}},\\
\mathcal L_{\mathrm{Stage\,2}}
&=\lambda_r\mathcal L_{\mathrm{risk}},
\end{aligned}
\label{eq:carm_training_objective}
\end{equation}
where $\lambda_e,\lambda_r$ are loss weights and $\lambda_{\mathrm{sat}}=0.01$.
Stage~1 sets $x_\theta^j\equiv0$ and
trains the encoder and both branches; Stage~2 freezes them and trains
only $g_r$.

\section{Experiments}
\label{sec:experiments}

\noindent\textbf{Setup.}
An instance contains one ego trajectory and surrounding agents over an
$H=4$\,s window ($K=20$ bins). Collision-agreement results use the held-out
DeepAccident~\cite{deepaccident} test split (train/val/test:
4967/1656/1656) with native, sensor-independent collision labels; the
fine-grained analysis pools all 8279 instances across the three splits.
Cross-domain results apply the DeepAccident-trained CARM zero-shot to all
3679 Bench2Drive~\cite{bench2drive} instances. As Bench2Drive lacks
native collision annotations, we use the geometric 3D-TTC overlap test
(Eq.~\eqref{eq:high_level_3d_ttc}); this is circular for methods
derived from the same test, as marked in the tables. Collision-agreement
claims therefore rely on in-domain native labels. CARM uses width 64 and
dropout 0.1 and is trained with Adam (learning rate $10^{-3}$, weight
decay $10^{-4}$, batch size 32) for 40 Stage-1 and 20 Stage-2 epochs;
all numbers use three seeds.

\noindent\textbf{Baselines and metrics.}
Rule-based baselines include the analytic target $y_{\mathrm{analytic}}$
(Eq.~\eqref{eq:analytic_risk_target}), raw 3D-TTC, PET~\cite{wang2021review}, RSS
\cite{shalevshwartz2017rss}, and Forward TTC, our instantiation of Li
\emph{et al.}~\cite{li2024beyond}, which projects each party under
held-constant control rather than reading the observed future. We report
its 2-D (constant turn rate and velocity) and 1-D (no turning) forms.
Learning-based baselines are Lu \emph{et al.}'s unsupervised
surrogate-safety embedding~\cite{lu2022learning}, retrained on
DeepAccident, and GSSM~\cite{jiao2026gssm} with its released
naturalistic-driving checkpoint. Collision agreement uses AUROC and
AUPRC against binary labels. Fine-grained discrimination on native-label
collision-free instances uses Spearman's $\rho$ against the trajectory-level
proxy $a_{\max}=\max_j\max_t a^j(t)$ and negative worst-case geometric
margin $-m$ (Eq.~\eqref{eq:future_safety_margin}), finite only when overlap
occurs within the horizon. Planner comparisons keep surrounding agents fixed
within each scenario.

\subsection{Consistency with Collision-Based Evaluation}
\label{sec:exp_consistency}

Table~\ref{tab:instance_consistency} reports AUROC and AUPRC against
DeepAccident's native collision label. FutureRisk reaches $0.793$/$0.490$,
outperforming all baselines on both metrics; the strongest are
Forward TTC (2-D) on AUROC ($0.685$) and Lu \emph{et al.} on AUPRC
($0.452$).

The rule-based baselines differ in what they read. $y_{\mathrm{analytic}}$
and raw 3D-TTC read the observed future through a discrete-frame box
overlap test, which can miss contacts recorded by the native label
(Sec.~\ref{sec:carm_training}); both AUROCs remain near $0.65$. Forward TTC is
slightly higher, plausibly because continuous projection is more
forgiving of near-contact than exact overlap; its 2-D and 1-D forms are
indistinguishable. PET scores similarly to $y_{\mathrm{analytic}}$
and raw 3D-TTC ($0.667$/$0.303$), consistent with all three reading
proximity at a single frame rather than a continuously projected
future. RSS is lowest ($0.529$/$0.199$): its fixed envelope
over current speed and gap is too conservative and scene-agnostic to
track interactions ending in contact. Among learning-based baselines,
Lu \emph{et al.}'s sequence embedding achieves the best baseline AUPRC
but a middling AUROC ($0.662$), while GSSM, applied zero-shot from a
naturalistic-driving checkpoint, falls below all TTC-based baselines and
above only RSS ($0.567$/$0.226$). FutureRisk's advantage comes from
structure absent from these baselines: a safety-time distribution
composed with an independently supervised critical-event distribution;
Sec.~\ref{sec:exp_tradeoff} attributes the gain to each term.

\begin{table}[t]
\footnotesize
\renewcommand{\arraystretch}{1.25}
\setlength{\tabcolsep}{4pt}
\centering
\caption{Instance-level discrimination against the binary collision label,
DeepAccident.}
\label{tab:instance_consistency}
\begin{tabular}{lcc}
\toprule
Method & AUROC $\uparrow$ & AUPRC $\uparrow$ \\
\midrule
\multicolumn{3}{l}{\textit{Rule-based}} \\
$y_{\mathrm{analytic}}$ & 0.648 & 0.372 \\
raw 3D-TTC & 0.646 & 0.350 \\
Forward TTC (2-D)~\cite{li2024beyond} & 0.685 & 0.425 \\
Forward TTC (1-D)~\cite{li2024beyond} & 0.684 & 0.424 \\
PET~\cite{wang2021review} & 0.667 & 0.303 \\
RSS~\cite{shalevshwartz2017rss} & 0.529 & 0.199 \\
\midrule
\multicolumn{3}{l}{\textit{Learning-based}} \\
Lu \emph{et al.}~\cite{lu2022learning} & 0.662 & 0.452 \\
GSSM~\cite{jiao2026gssm} & 0.567 & 0.226 \\
\midrule
\multicolumn{3}{l}{\textit{Our method}} \\
FutureRisk (CARM) & \textbf{0.793} & \textbf{0.490} \\
\bottomrule
\end{tabular}
\end{table}

\subsection{Fine-Grained Discrimination}
\label{sec:exp_finegrained}

\begin{figure}[t]
\centering
\includegraphics[width=0.8\columnwidth]{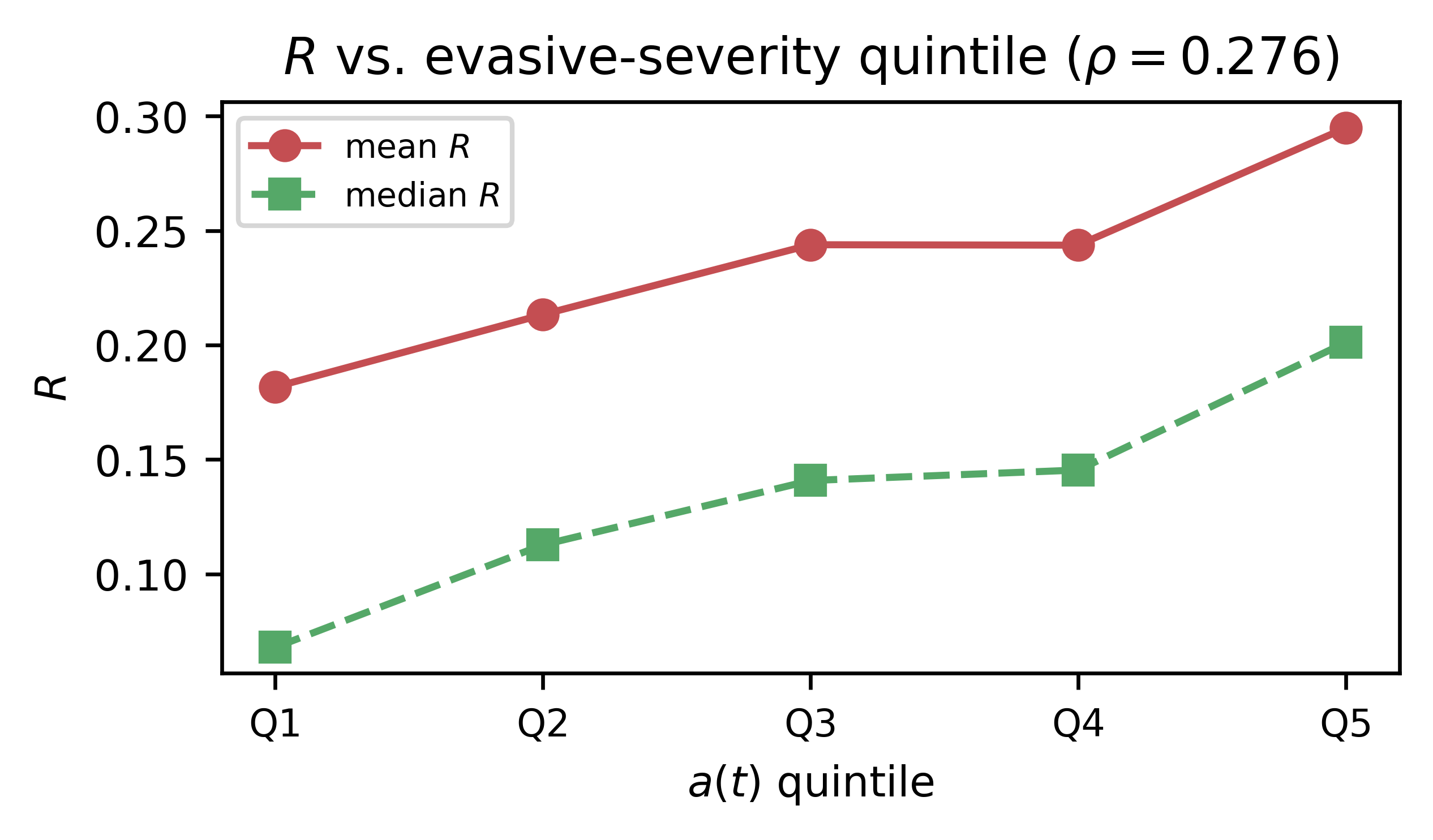}
\caption{Mean and median $R$ by quintile of the evasive-severity proxy
$a_{\max}$ on pooled native-label collision-free DeepAccident instances
($n=6630$). Both show an overall increasing trend with severity; Spearman $\rho=0.276$.}
\label{fig:severity_bins}
\vspace{-2mm}
\end{figure}

Table~\ref{tab:finegrained_baselines} evaluates whether each score can order native-label collision-free trajectories using two complementary references. The evasive-severity proxy $a_{\max}$ is related to the signal defining critical-event labels, but supervision provides only threshold-crossing time, not continuous magnitude. Thus, $a_{\max}$ tests graded ordering beyond thresholded supervision, while $-m$ provides a geometry-based reference. FutureRisk correlates with both: $\rho=0.276$ against $a_{\max}$ ($n=6630$) and $\rho=0.290$ against $-m$ ($n=286$). Fig.~\ref{fig:severity_bins} further shows an overall increasing trend in $R$ across $a_{\max}$ quintiles, including those well below the $95^{\mathrm{th}}$-percentile critical-event threshold (Eq.~\eqref{eq:carm_critical_event}).

The two references favor different hand-crafted quantities by construction. Since $a_{\max}$ depends on accelerations, speed-and-gap measures correlate strongly with it (RSS $0.363$, Forward TTC $0.274$), while those reading the already-evaded future are near zero ($y_{\mathrm{analytic}}$ $0.007$, raw 3D-TTC $0.006$). Where $m$ exists, $y_{\mathrm{analytic}}=\phi(m)$ is circular, while raw 3D-TTC and Forward TTC collapse to one value. On the other reference, RSS falls to $0.122$ against $-m$; Lu \emph{et al.}'s embedding similarly splits ($0.430$ against $-m$, $-0.001$ against $a_{\max}$). FutureRisk is the only score above $0.25$ on both references while also leading collision agreement (Sec.~\ref{sec:exp_consistency}).

\begin{table}[t]
\scriptsize
\renewcommand{\arraystretch}{0.92}
\setlength{\tabcolsep}{2.2pt}
\centering
\caption{Fine-grained discrimination on collision-free trajectories.
In-domain: $n=6630/286$ for $a(t)/m$; cross-domain: $n=3333$ zero-shot.
Cross-domain $m$ is undefined ($m=+\infty$ throughout).}
\label{tab:finegrained_baselines}

\resizebox{0.9\columnwidth}{!}{
\begin{tabular}{lcccc}
\toprule
& \multicolumn{2}{c}{$\rho$ vs.\ $a(t)$}
& \multicolumn{2}{c}{$\rho$ vs.\ $m$} \\
\cmidrule(lr){2-3}\cmidrule(lr){4-5}
Method & in & cross & in & cross \\
\midrule

\multicolumn{5}{l}{\textit{Rule-based}} \\
$y_{\mathrm{analytic}}$
    & 0.007 & -- \emph{(const.)} & 1.000 \emph{(circ.)} & n/a \\
raw 3D-TTC
    & 0.006 & -- \emph{(const.)} & -- \emph{(const.)} & n/a \\
Forward TTC (2-D)~\cite{li2024beyond}
    & 0.274 & 0.283 & -- \emph{(const.)} & n/a \\
Forward TTC (1-D)
    & 0.265 & 0.291 & -- \emph{(const.)} & n/a \\
PET
    & $-0.093$ & 0.167 & 0.155 & n/a \\
RSS~\cite{shalevshwartz2017rss}
    & \textbf{0.363} & 0.295 & 0.122 & n/a \\

\midrule
\multicolumn{5}{l}{\textit{Learning-based}} \\
Lu \emph{et al.}~\cite{lu2022learning}
    & $-0.001$ & 0.110 & 0.230 & n/a \\
GSSM~\cite{jiao2026gssm}
    & 0.070 & 0.317 & 0.080 \emph{(n.s.)} & n/a \\

\midrule
\multicolumn{5}{l}{\textit{Our method}} \\
FutureRisk (CARM)
    & 0.276 & \textbf{0.436} & \textbf{0.290} & n/a \\

\bottomrule
\end{tabular}}
\end{table}

\subsection{Interpretability--Adaptivity Trade-off}
\label{sec:exp_tradeoff}

We define aggregate anchoring as $100\%$ for RSS, $0\%$ for Lu
\emph{et al.}, and
$1-\mathbb E[|x_\theta|/|\operatorname{logit}\widetilde
R_{\mathrm{CARM}}|]$ for FutureRisk, averaged per agent over the
DeepAccident test set. The first two follow from model structure; GSSM
has no closed-form/free decomposition and is marked n/a. As the
diagnostic is unclipped, values may fall outside $[0,1]$.

Table~\ref{tab:interpretability_tradeoff} shows that FutureRisk retains
$70.1\%$ anchoring while leading collision agreement in-domain
($0.793$) and cross-domain ($0.827$). Lu \emph{et al.}'s cross-domain
$0.990$ is circular because its TTC input defines the label. RSS is fully
anchored but has low in-domain AUROC ($0.529$), while Lu \emph{et al.}
is fully free but fails to track evasive severity ($-0.001$). GSSM
remains below FutureRisk in AUROC ($0.567$/$0.817$) and correlation
($0.070$).

\begin{figure}[t]
\centering
\IfFileExists{figures/anchored_fraction_distribution.drawio.png}{%
\includegraphics[width=0.85\columnwidth]{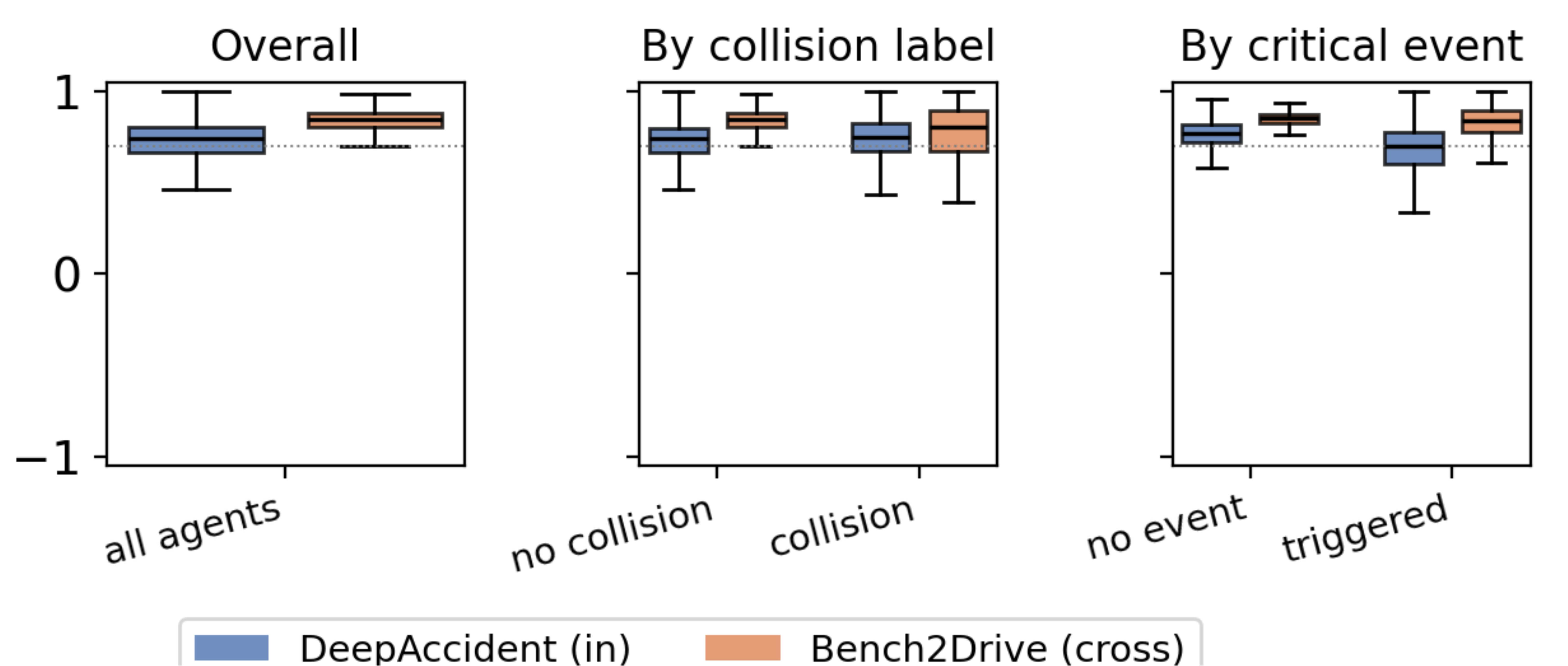}%
}{\vspace{1.6in}}
\caption{Per-agent anchored-fraction distributions on DeepAccident and
Bench2Drive, grouped by collision and critical-event labels. The dotted
line marks the $70.1\%$ in-domain mean; the display is clipped to
$[-1,1]$, while all values remain in the reported statistics.}
\label{fig:anchored_fraction_distribution}
\vspace{-1mm}
\end{figure}

Across $75{,}752$ in-domain agents,
Fig.~\ref{fig:anchored_fraction_distribution} shows a median anchored
fraction of $0.741$, versus $0.846$ cross-domain. Critical events include
collisions and evasive maneuvers, capturing non-collision near misses.
A small unstable tail occurs when
$\operatorname{logit}\widetilde R_{\mathrm{CARM}}\approx0$; excluding it
raises the in-domain mean from $0.701$ to $0.720$ without changing the
median, and affects only $0.21\%$/$0.19\%$ of in-/cross-domain agents.
Anchoring separates critical events more than collisions: its median is
$0.700$ for finite ground-truth $T_{\mathrm{evt}}$ and $0.769$ otherwise
(rank-biserial $r=0.371$), whereas collision status changes it by only
$0.009$ ($r=-0.066$). Thus, the residual is relatively more influential
in critical interactions, where the physical prior is under greater
strain.

Table~\ref{tab:residual_ablation} adds safety-time and residual terms to
an always-trained critical-event branch. The uncorrected
$\tau_{\mathrm{phy}}$ reaches $0.731$/$0.799$ in-/cross-domain AUROC.
Adding $\Delta_\theta$ raises in-domain AUROC by $0.022$;
$\sigma_s$ helps jointly (Stage-1, $0.764$) but not alone ($0.718$).
The Stage-2 residual gives the largest increment
($0.764\to0.793$) and the best result in all four columns.
From the raw prior $\to$ Stage-1 $\to$ full CARM, three columns improve
monotonically; cross-domain AUPRC dips from $0.344$ to $0.333$ before
recovering to $0.354$. Bounds on $\Delta_{\max}$ and $\beta$ keep these
gains anchored to the prior.

Finally, we isolate independent critical-event supervision by retraining
Full CARM with $\lambda_e=0$ while retaining all components.
Table~\ref{tab:event_ablation} shows that removing
$\mathcal{L}_{\mathrm{evt}}$ slightly improves in-domain collision AUROC
($0.821$ vs.\ $0.793$) while leaving cross-domain performance similar
($0.829$ vs.\ $0.827$), but substantially reduces correlation with
$a_{\max}$ ($0.276\to0.082$). Thus, critical-event supervision
contributes primarily to fine-grained severity discrimination among
collision-free interactions rather than binary collision agreement.

\begin{table}[t]
\centering
\caption{Ablation of independent critical-event supervision. The
event head is retained when removing $\mathcal{L}_{\mathrm{evt}}$.}
\label{tab:event_ablation}
\setlength{\tabcolsep}{3pt}
\begin{tabular*}{\columnwidth}{@{\extracolsep{\fill}}lccccc@{}}
\toprule
& AUROC & AUPRC & AUROC & AUPRC & $\rho$ \\
Model
& (in) $\uparrow$
& (in) $\uparrow$
& (cross) $\uparrow$
& (cross) $\uparrow$
& $(R, a_{\max})$ $\uparrow$ \\
\midrule
w/o $\mathcal{L}_{\mathrm{evt}}$
& \textbf{0.821} & \textbf{0.502} & \textbf{0.829} & \textbf{0.357} & 0.082 \\
Full CARM
& 0.793 & 0.490 & 0.827 & 0.354 & \textbf{0.276} \\
\bottomrule
\end{tabular*}
\end{table}

\begin{table}[t]
\footnotesize
\renewcommand{\arraystretch}{1.25}
\setlength{\tabcolsep}{3pt}
\centering
\caption{Ablation of safety-time and residual terms; the learned
critical-event branch is retained throughout.}
\label{tab:residual_ablation}
\begin{tabular}{lcccc}
\toprule
Model & \tabstack{AUROC}{(in) $\uparrow$} & \tabstack{AUPRC}{(in) $\uparrow$} & \tabstack{AUROC}{(cross) $\uparrow$} & \tabstack{AUPRC}{(cross) $\uparrow$} \\
\midrule
raw safety-time prior & 0.731 & 0.442 & 0.799 & 0.344 \\
$+\,\Delta_\theta$ only & 0.753 & 0.483 & 0.812 & 0.341 \\
$+\,\sigma_s$ only & 0.718 & 0.417 & 0.802 & 0.331 \\
Stage-1 only (both, $g_r\equiv0$) & 0.764 & 0.459 & 0.818 & 0.333 \\
Full CARM & \textbf{0.793} & \textbf{0.490} & \textbf{0.827} & \textbf{0.354} \\
\bottomrule
\end{tabular}
\end{table}

\begin{table}[t]
\footnotesize
\renewcommand{\arraystretch}{1.25}
\setlength{\tabcolsep}{2.5pt}
\centering
\caption{Interpretability--adaptivity trade-off. Anchoring is structural
for RSS and Lu \emph{et al.}, empirical for FutureRisk, and undefined for
GSSM's non-decomposable output.}
\label{tab:interpretability_tradeoff}
\begin{tabular}{lcccc}
\toprule
Method & Anchored $\uparrow$ & \tabstack{AUROC}{(in) $\uparrow$} & \tabstack{AUROC}{(cross) $\uparrow$} & $\rho$ vs.\ $a_{\max}$ $\uparrow$ \\
\midrule
RSS~\cite{shalevshwartz2017rss} & $100\%$ & 0.529 & 0.662 & \textbf{0.363} \\
Lu \emph{et al.}~\cite{lu2022learning} & $0\%$ & 0.662 &  0.990~\emph{(circ.)} & $-$0.001 \\
GSSM~\cite{jiao2026gssm} & n/a & 0.567 & 0.817 & 0.070 \\
FutureRisk (CARM) & $70.1\%$ & \textbf{0.793} & \textbf{0.827} & 0.276 \\
\bottomrule
\end{tabular}
\end{table}

\subsection{Planner Comparison and Qualitative Analysis}
\label{sec:exp_planner_screening}

We test whether FutureRisk ranks planners consistently with collision
rate but at finer resolution. We use four released Bench2Drive
planners~\cite{bench2drivezoo}: AD-MLP (kinematics only), TCP (single
front camera), and UniAD-Tiny and VAD-Base (surround-view BEV), using
their eligible \emph{AccidentTwoWays} clips ($n=13$ for TCP and $n=15$
otherwise). Surrounding agents remain fixed, while the ego box follows
the predicted path, interpolated onto CARM's time grid and extrapolated
at constant velocity beyond its horizon. All trajectories are scored
zero-shot by the DeepAccident-trained checkpoint. Collision rate is the
fraction with geometric overlap (Eq.~\eqref{eq:high_level_3d_ttc}).

\begin{figure}[t]
\vspace{-1mm}
\centering
\includegraphics[width=2.4in]{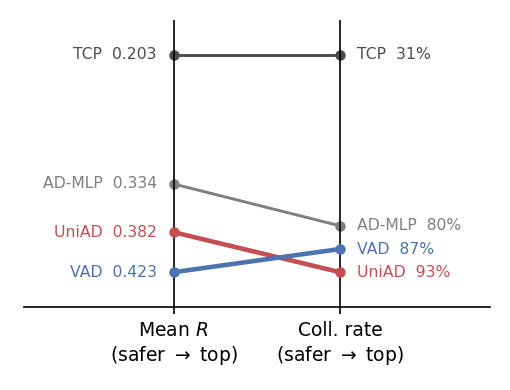}
\caption{Planner ranking by mean $R$ (left) and geometric collision rate
(right) over available clips ($n=13$ for TCP; $n=15$ otherwise); safer planners are plotted higher.
Parallel lines (TCP, AD-MLP) indicate agreement; crossing lines (UniAD,
VAD) indicate disagreement.}
\label{fig:planner_slope_chart}
\vspace{-2.5mm}
\end{figure}

\begin{figure*}[t]
\vspace{-2mm}
\centering
\includegraphics[width=0.95\textwidth]{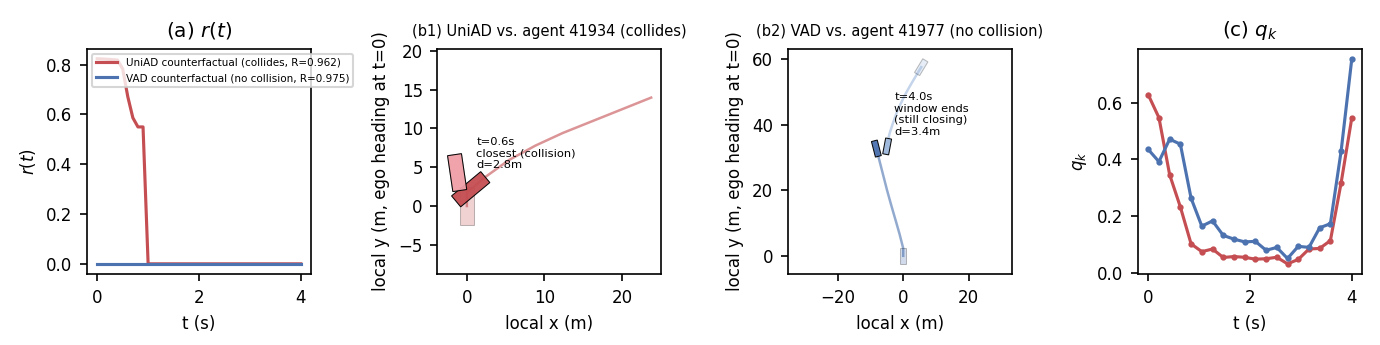}
\caption{One AccidentTwoWays scenario (Route~1105) under UniAD (red) and VAD (blue). UniAD collides at $t=0.6$\,s ($R=0.962$), while VAD avoids overlap within 4\,s but closes from $57.8$\,m to $3.4$\,m at about $4$\,m/s ($R=0.975$). (a) Closed-form risk $r(t)$: nonzero for UniAD but zero for VAD. (b1)/(b2) Bird's-eye trajectories; light boxes mark $t=0$ and solid boxes the moment of interest. (c) Critical-event hazard $q_k$ for the relevant agent.}
\label{fig:case_study_uniad_vs_vad}
 \vspace{-2mm}
\end{figure*}

Fig.~\ref{fig:planner_slope_chart} compares mean $R$ and collision rate.
Both rank TCP safest and AD-MLP second, but reverse UniAD and VAD: mean
$R$ favors UniAD, whereas collision rate favors VAD. This gap
($0.382$ vs.\ $0.423$) arises almost entirely from one clip in
Fig.~\ref{fig:case_study_uniad_vs_vad}. VAD never overlaps an agent
within 4\,s, so the overlap test and closed-form margin register nothing
($m(t)=+\infty$, $r(t)\equiv0$, panel~(a)). Yet its distance to one agent
falls from $57.8$\,m to $3.4$\,m while closing at about $4$\,m/s
(panel~(b2)), increasing the critical-event hazard $q_k$ (panel~(c))
and yielding $R=0.975$. UniAD instead collides at $t=0.6$\,s
(panel~(b1)) and scores $R=0.962$. Excluding this clip gives VAD mean
$R=0.383$, nearly matching UniAD's $0.382$. This disagreement reflects
an interaction the fixed-horizon overlap test misses before contact:
without overlap, the geometric target $y$ is $0$, while the
independently supervised hazard $q_k$ remains informative.

\begin{figure}[t]
% \vspace{2mm}
\centering
\includegraphics[width=\columnwidth]{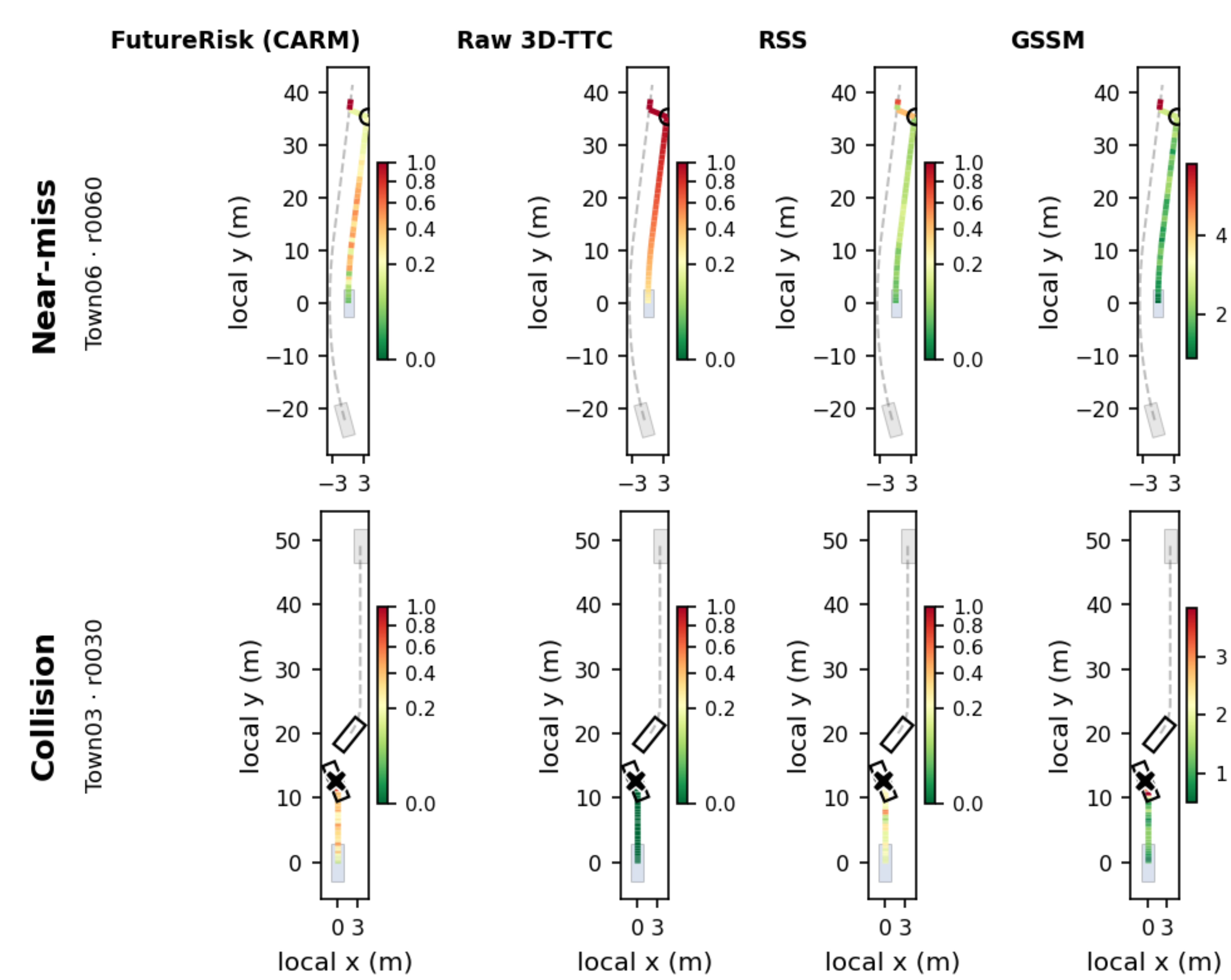}
\caption{Per-frame risk for two DeepAccident instances (rows) under four
methods (columns): a near-miss (top; native label $0$) and a collision
missed by geometric overlap (bottom; native label $1$, contact at
$t=4.0$\,s). Path color indicates risk (green low, red high); GSSM is
normalized to its own range per row, while the others share one $[0,1]$
scale. Gray dashed: interacting agent; \textsf{X}: contact; \textsf{o}:
closest approach.}
\label{fig:qualitative_per_timestep_comparison}
\vspace{-3mm}
\end{figure}

Fig.~\ref{fig:qualitative_per_timestep_comparison} shows per-frame scores
for two DeepAccident instances: a near-miss (top) and a collision missed
by geometric overlap (bottom; native contact at $t=4.0$\,s). For the
missed collision, raw 3D-TTC remains $0.0$ across all $41$ frames. RSS
and GSSM peak at $0.552$ at $t=3.1$\,s and $3.85$ at $t=3.6$\,s,
respectively (the latter in its own unbounded units), while FutureRisk
is elevated at confirmed contact ($R=0.478$). Its $[0,1]$ scale remains
anchored to the physics prior (Sec.~\ref{sec:exp_tradeoff}), unlike
GSSM's unbounded intensity or RSS's collision-insensitive envelope.
On the near-miss, all four methods rise toward closest approach without
indicating collision; FutureRisk spikes at $t=0.3$\,s and settles within
one frame.

\subsection{Perturbation Sweep and Hyperparameter Sensitivity}
\label{sec:exp_sensitivity}

\begin{figure}[t]
\vspace{2mm}
\centering
\includegraphics[width=0.8\columnwidth]{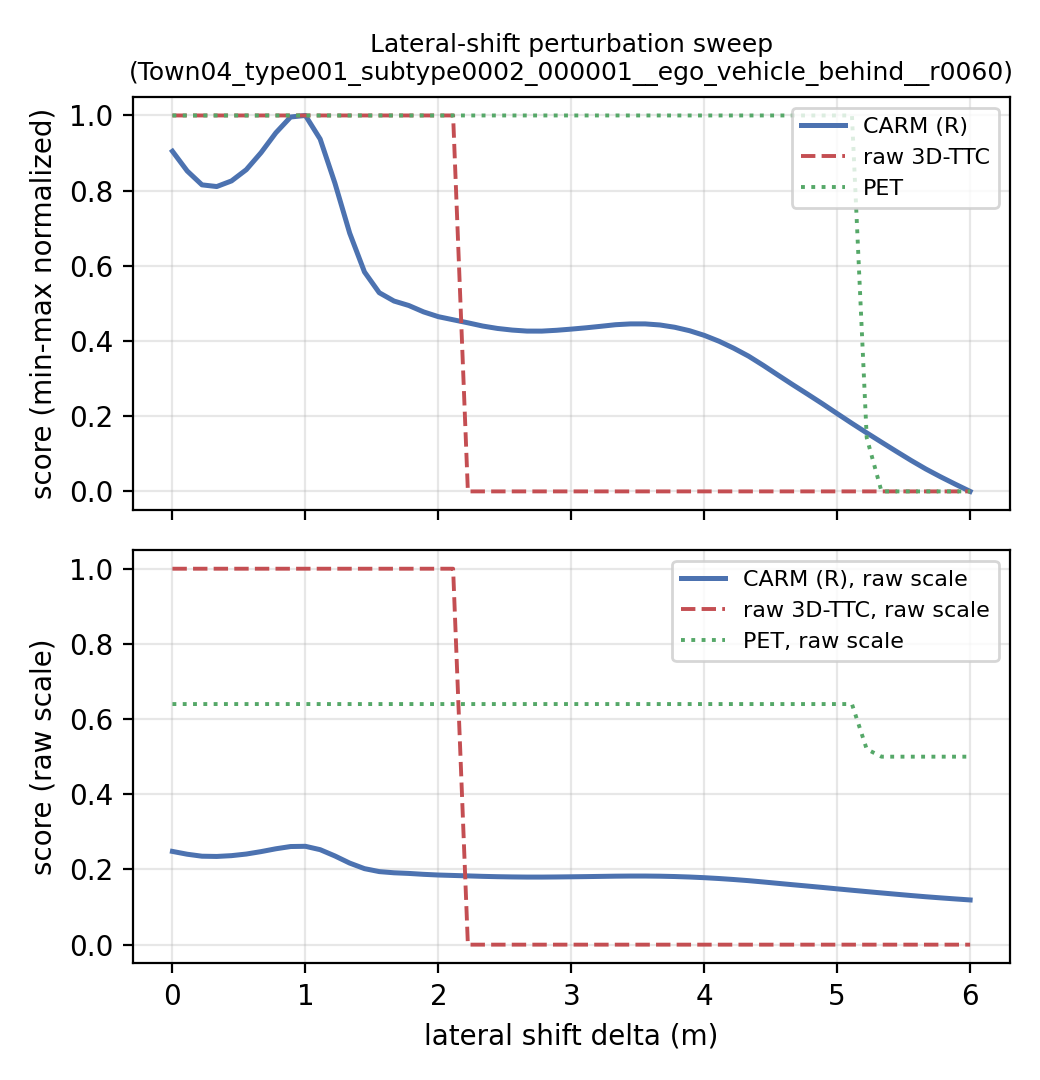}
\caption{$R$, 3D-TTC score, and PET score as the ego path of one
DeepAccident collision instance is shifted laterally by $\delta$ (top:
min--max normalized; bottom: raw scores). The rule-based curves change
abruptly near their thresholds, while $R$ changes smoothly.}
\label{fig:perturbation_sweep}
\vspace{-2mm}
\end{figure}

\noindent\textbf{Perturbation sweep.}
Starting from a real DeepAccident collision, we shift the ego path
laterally from $\delta=0$ (collision) to $\delta=6$\,m (ample clearance),
sampling $55$ offsets and recomputing $R$, 3D-TTC, and PET
(Fig.~\ref{fig:perturbation_sweep}). The rule-based scores change
abruptly: 3D-TTC drops from $1.0$ to $0.0$ between adjacent offsets
$0.02$\,m apart near the overlap boundary ($\delta\approx2.15$\,m),
while normalized PET changes over $\delta\approx5.1$--$5.4$\,m.
In contrast, $R$ changes gradually across the range, with a maximum
adjacent change of $0.019$ and total variation of $0.19$, demonstrating
smoother responses to small trajectory perturbations than the rule-based
scores.

\noindent\textbf{Hyperparameter sensitivity.}
The transition width $T_m$ in Eq.~\eqref{eq:continuous_risk}
monotonically rescales the margin and therefore preserves the ranking
of $y_{\mathrm{analytic}}$; across all $8279$ DeepAccident instances,
Kendall's $\tau=1$ between $T_m\in\{0.5,2.0\}$ and $T_m=1.0$.
Retraining CARM over $T_m\in\{0.5,1.0,2.0\}$ and critical-event
percentiles $\{90,95,99\}$, using $12$ Stage-1 and $6$ Stage-2 epochs,
yields in-domain AUROC within $0.735$--$0.754$ across all nine settings
(a maximum variation of $0.019$).

% \subsection{Binned Severity, Perturbation Sweep, and Hyperparameter Sensitivity}

% \textbf{Parametric perturbation sweep} (Fig.~\ref{fig:perturbation_sweep}):
% from a real DeepAccident collision, we shift ego's path laterally
% (perpendicular to per-frame heading) from $\delta=0$ (colliding) past
% the collision boundary to $\delta=6$m (clearly safe), $55$ steps,
% recomputing $R$, raw 3D-TTC, and PET at each. Raw 3D-TTC steps from
% $1.0$ to $0.0$ within a single $0.02$m step at the boundary
% ($\delta\approx2.15$m; total variation $=1.0$, the full scale); PET
% jumps similarly, later ($\delta\approx5.1$--$5.4$m); $R$ instead
% decreases smoothly (max step change $=0.019$, total variation $=0.19$).

\section{Conclusion}

FutureRisk complements binary collision-based evaluation by continuously
and interpretably distinguishing trajectories with identical collision
outcomes but different latent risks. Across experiments, it agrees
strongly with observed collisions while better discriminating
collision-free interactions. By combining a physics-based safety prior
with learned adaptation, it captures risky interactions overlooked by
fixed geometric criteria while retaining physical interpretability, with
roughly $70\%$ of each score still attributable to the prior, a
combination no existing surrogate measure provides.

\noindent\textbf{Limitation and Future Work.} Fine-grained risk has no
unique ground truth, so our evaluation relies on indirect references
rather than direct annotation. More broadly, bounded anchoring suggests a
general way to build safety criteria: learning may reshape a physical
quantity, but only within a stated bound. Future work will apply this
principle to online risk estimation and to criteria beyond time-based
margins.

% Corrected reference list for root.tex, verified 2026-09-14 against Crossref / publisher / arXiv records.
% Drop-in replacement for everything from \begin{thebibliography}{99} to the first \end{thebibliography}.
% - Numbered in order of first citation (IEEE); duplicate jiao2026gssm merged into one entry (20 entries).
% - Only verified values; no DOIs or months (optional in IEEE style), "Art. no." for article-numbered journals,
%   "in Proc." for conference papers, IEEE arXiv form, first author + et al. for more than six authors.
% - No \cite key changes: the body text compiles unchanged.
% - Keeps the paper's \scriptsize: with the template default (\footnotesize) the paper runs to 9 pages (tested).
% Tested in the 2026-09-14 icra.zip: 8 pages, 0 errors, 0 overfull boxes, citations print [1]..[20] in reading order.

\end{document}